\documentclass[sigconf,nonacm]{acmart}

\usepackage{bm} 

\usepackage{multirow}
\usepackage{soul}  
\usepackage{balance}  

\AtBeginDocument{%
  }

\begin{document}

\title{Spatiotemporal Degradation-Aware 3D Gaussian Splatting for Realistic Underwater Scene Reconstruction}
\thanks{This is the author version of the paper published in Proceedings of ACM Multimedia 2025. The version of record is available at \url{https://doi.org/10.1145/3746027.3754888}.}


\author{Shaohua Liu}
\orcid{0009-0005-9280-4667}
\affiliation{%
  \department{School of Astronautics}
  \institution{Beihang University}
  \city{Beijing}
  \country{China}}
\additionalaffiliation{%
  \department{Shen Yuan Honors College}
  \institution{Beihang University}
  \city{Beijing}
  \country{China}
}  
\email{liushaohua@buaa.edu.cn}

\author{Ning Gao}
\orcid{0009-0004-0451-281X}
\affiliation{%
  \department{School of Astronautics}
  \institution{Beihang University}
  \city{Beijing}
  \country{China}}
\email{gaoning_ai@buaa.edu.cn}

\author{Zuoya Gu}
\orcid{0009-0000-7575-9247}
\affiliation{%
  \department{School of Astronautics}
  \institution{Beihang University}
  \city{Beijing}
  \country{China}}
\email{zuoyagu@buaa.edu.cn}

\author{Hongkun Dou}
\orcid{0000-0001-6185-5369}
\affiliation{%
  \department{School of Astronautics}
  \institution{Beihang University}
  \city{Beijing}
  \country{China}}
\email{douhk@buaa.edu.cn}

\author{Yue Deng}
\orcid{0000-0003-2871-8922}
\affiliation{%
  \department{School of Artificial Intelligence}  
  \institution{Beihang University}
  \city{Beijing}
  \country{China}}
\affiliation{%
  \institution{Zhongguancun Academy}
  \city{Beijing}
  \country{China}}
\email{ydeng@buaa.edu.cn}

\author{Hongjue Li}
\orcid{0000-0002-0504-2555}
\authornote{Corresponding author.}
\affiliation{%
  \institution{Beihang University}
  \department{School of Astronautics}
  \city{Beijing}
  \country{China}
}
\additionalaffiliation{%
  \department{State Key Laboratory of High-Efficiency Reusable Aerospace Transportation Technology}  
  \institution{Beihang University}
  \city{Beijing}
  \country{China}
}
\additionalaffiliation{%
  \department{Beijing Key Laboratory of System Design for Reusable Launch Vehicle}
  \institution{Beihang University}
  \city{Beijing}
  \country{China}
}
\email{lihongjue@buaa.edu.cn}

\renewcommand{\shortauthors}{Shaohua Liu et al.}

\begin{abstract}

Reconstructing realistic underwater scenes from underwater video remains a meaningful yet challenging task in the multimedia domain. The inherent spatiotemporal degradations in underwater imaging, including caustics, flickering, attenuation, and backscattering, frequently result in inaccurate geometry and appearance in existing 3D reconstruction methods. While a few recent works have explored underwater degradation-aware reconstruction, they often address either spatial or temporal degradation alone, falling short in more real-world underwater scenarios where both types of degradation occur. We propose MarineSTD-GS, a novel 3D Gaussian Splatting-based framework that explicitly models both temporal and spatial degradations for realistic underwater scene reconstruction. Specifically, we introduce two paired Gaussian primitives: Intrinsic Gaussians represent the true scene, while Degraded Gaussians render the degraded observations. The color of each Degraded Gaussian is physically derived from its paired Intrinsic Gaussian via a Spatiotemporal Degradation Modeling (SDM) module, enabling self-supervised disentanglement of realistic appearance from degraded images. To ensure stable training and accurate geometry, we further propose a Depth-Guided Geometry Loss and a Multi-Stage Optimization strategy. We also construct a simulated benchmark with diverse spatial and temporal degradations and ground-truth appearances for comprehensive evaluation. Experiments on both simulated and real-world datasets show that MarineSTD-GS robustly handles spatiotemporal degradations and outperforms existing methods in novel view synthesis with realistic, water-free scene appearances.

\end{abstract}



\begin{teaserfigure}
  \includegraphics[width=\textwidth]{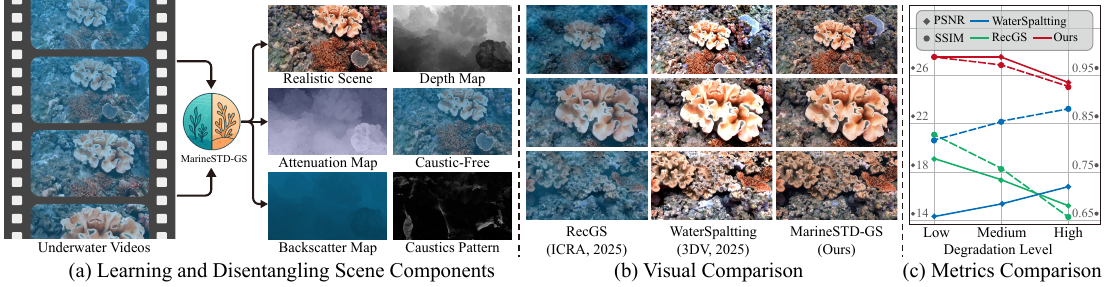}
\caption{(a) Given underwater video sequences with complex spatiotemporal degradations, our MarineSTD-GS performs holistic scene understanding by disentangling realistic scene representations—recovering true color and geometry (depth), and estimating spatial degradation parameters (e.g., attenuation and backscatter) as well as transient caustic patterns. (b) Compared to state-of-the-art underwater 3D reconstruction methods, our approach produces more faithful and consistent scene appearances, effectively reducing color distortion and caustic-induced flickering. (c) Quantitative results show that MarineSTD-GS consistently outperforms all baselines across varying degradation levels, demonstrating superior robustness. For more visual Comparison and Results, please refer to our project page via \url{https://marinestd-gs.github.io/}.
}
  \Description{The figure shows: (a) our model MarineSTD-GS processing underwater video frames to produce realistic scene, depth, attenuation, backscatter, and caustic maps; (b) compares our results with other underwater 3D reconstruction methods, highlighting reduced color distortion and flickering; (c) a line chart comparing PSNR and SSIM across various degradation levels, where our method consistently achieves superior results.}
  \label{fig:teaser}
\end{teaserfigure}


\maketitle

\section{Introduction}

Reconstructing realistic scenes from underwater video is critical for a wide range of marine multimedia applications, including underwater archaeology~\cite{liu2022twin, zhou2024hifi}, ecological monitoring~\cite{xue2023learning, yuval2024releasing}, and immersive virtual reality~\cite{johnson2017high, llorach2023experience}. However, underwater imaging is inherently affected by a range of degradation factors, which can be broadly divided into spatial and temporal categories. Spatial degradations such as distance-dependent attenuation and backscattering often cause color distortions and haze effects~\cite{boittiaux2024sucre}. On the other hand, temporal degradations including caustics and flickering lead to inconsistent lighting and severe local brightness fluctuations~\cite{forbes2018deepcaustics, sauder2024self,agrafiotis2018underwater}. These spatiotemporal degradations violate the multi-view consistency assumption required by most advanced 3D reconstruction methods such as Neural Radiance Field (NeRF)\cite{mildenhall2021nerf} and 3D Gaussian Splatting (3DGS)~\cite{kerbl20233d}, resulting in inaccurate geometry and appearance, and ultimately hindering downstream applications~\cite{fei20243d}.

Recent efforts~\cite{levy2023seathru, li2025watersplatting, yang2024seasplat, zhang2024recgs} in underwater-specific 3D reconstruction have attempted to address certain types of degradation. SeaThruNeRF~\cite{levy2023seathru}, WaterSplatting~\cite{li2025watersplatting}, and SeaSplat~\cite{yang2024seasplat} introduce attenuation and scattering models to decouple realistic scene appearance from spatially degraded underwater observations. RecGS~\cite{zhang2024recgs}, on the other hand, incorporates low-pass filtering and recurrent updates into the 3DGS pipeline to mitigate flickering artifacts. However, since these methods typically address only one aspect of degradation, they struggle to recover realistic geometry and appearance when both spatial and temporal degradations occur, as commonly seen in real underwater environments.

To address this challenge, we propose MarineSTD-GS, a novel 3D Gaussian-based framework that explicitly models both spatial and temporal degradation in underwater imaging and disentangles realistic scene representations from degraded videos through self-supervised learning. Specifically, we employ two paired types of Gaussian primitives: Intrinsic Gaussians represent the underlying realistic scene, while Degraded Gaussians are used to render degraded observations. Each pair of Intrinsic and Degraded Gaussians shares the same geometric attributes—including position, covariance, and opacity—while the color of the Degraded Gaussian is derived from its intrinsic counterpart via a physically grounded Spatiotemporal Degradation Modeling (SDM) module. To ensure robust geometry learning under degraded visual cues, we incorporate a Depth-Guided Geometry Loss that leverages monocular depth priors, and adopt a Multi-Stage Optimization strategy to stabilize training and refine appearance under coupled spatiotemporal degradations. To enable comprehensive evaluation, we also construct a simulated underwater dataset that includes diverse scenes, distinct temporal degradation patterns (e.g., caustics), and spatial degradation levels characterized by different attenuation and backscattering settings.

Our contributions can be summarized as follows:

\begin{itemize}
    \item We present MarineSTD-GS, the first 3DGS-based framework that explicitly models both spatial and temporal degradations in underwater imaging, enabling the disentangled reconstruction of realistic scene content and water-related degradation parameters.
    
    \item We develop a Multi-Stage Optimization strategy for stable training and a Depth-Guided Geometry Loss to enhance both global structure and local geometric fidelity.
        
    \item We construct and release a comprehensive simulated underwater benchmark with diverse scenes and varied temporal and spatial degradations for standardized evaluation.
    
    \item Extensive experiments on both simulated and real-world scenes show that MarineSTD-GS achieves state-of-the-art performance in color accuracy, flickering suppression, and visibility restoration.
\end{itemize}

\begin{figure*}[h]
  \centering
  \includegraphics[width=0.95\linewidth]{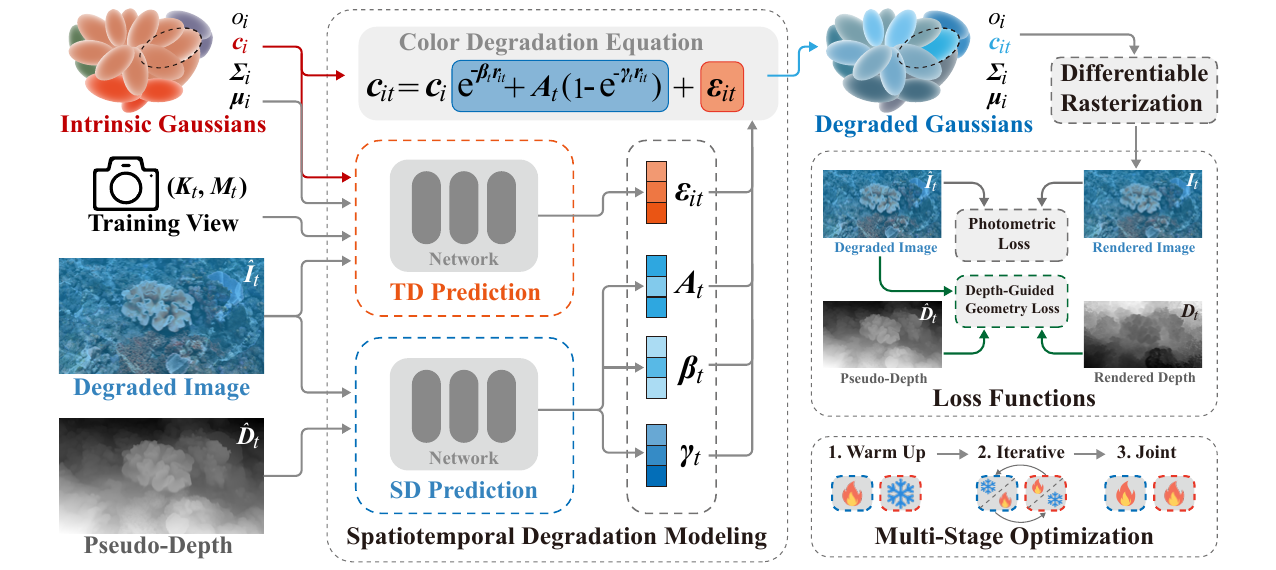}
  \caption{ Pipeline of MarineSTD-GS. 
   Given a training view at a specific time, the Spatiotemporal Degradation Modeling (SDM) module predicts the color degradation parameters for each pair of Intrinsic and Degraded Gaussians under the current spatiotemporal condition, and derives the degraded colors from their Intrinsic counterparts for rendering the corresponding underwater image and depth map. During optimization, in addition to photometric losses for reconstructing the degraded input, a Depth-Guided Geometry loss is employed to enhance geometric consistency. A Multi-Stage Optimization strategy is also adopted to ensure the stable training of the SDM components. 
} 
  \Description{A diagram illustrating the MarineSTD-GS pipeline, including input view, SDM module predicting degradation, rendering process, and optimization losses.}
\label{fig:pipeline}
\end{figure*}

\section{Related Work}

\subsection{3D Gaussian Splatting}
\label{sec:Related_3DGS}

3D Gaussian Splatting~\cite{kerbl20233d} represents scenes using a set of learnable 3D Gaussian primitives $\mathcal{G} = \{ \bm{\mu}, \bm{\Sigma}, o, \bm{c} \}$, where $\bm{\mu}$, $\bm{\Sigma}$, and $o$ are geometric attributes representing position, covariance, and opacity, respectively, while $\bm{c}$ denotes the color attribute, modeled via view-dependent spherical harmonics (SH). Given specific viewpoints, 3DGS employs a differentiable rasterization pipeline to render corresponding images, which are supervised by photometric losses (e.g., L1, D-SSIM) to optimize its learnable Gaussian primitives for accurate scene representation. We refer readers to the original paper~\cite{kerbl20233d} for more technical details of 3DGS.

Owing to its high fidelity and efficiency, 3DGS has been adapted to a variety of settings. These include sparse-view reconstruction~\cite{xiong2023sparsegs, li2024dngaussian, zhu2025fsgs}, large-scale scenes~\cite{lin2024vastgaussian, liu2024citygaussian}, dynamic scenes~\cite{yang2024deformable, wu20244d}, and challenging observations such as blurry images~\cite{lee2024deblurring, chen2024deblur} or in-the-wild image collections~\cite{zhang2024gaussian, dahmani2024swag, xu2024wild, xu2024splatfacto}. Recent works have also introduced geometric priors such as monocular depth~\cite{xiong2023sparsegs, li2024dngaussian}, or have applied various regularization strategies targeting edges~\cite{gong2024eggs}, frequencies~\cite{zhang2024fregs}, rank~\cite{hyung2024effective}, and topology~\cite{shen2025topology} to improve reconstruction quality. In addition, several methods~\cite{zhang2024pixel, fang2024mini, ye2024absgs} address the oversplatting problem in 3DGS optimization. In the context of underwater scene reconstruction, recent approaches~\cite{li2025watersplatting, yang2024seasplat, zhang2024recgs} integrate degradation modeling into the 3DGS framework via imaging equations or volumetric rendering. In contrast, MarineSTD-GS introduces a dual-Gaussian design and an SDM module to explicitly model both spatial and temporal degradations in a unified manner, leading to improved robustness and fidelity in underwater reconstruction.

\subsection{Underwater Restoration and Reconstruction}
\label{sec:URR}

Existing underwater image restoration methods can be broadly categorized into two branches: spatial degradation removal and temporal variation suppression. For spatial degradations, early methods rely on hand-crafted image priors~\cite{drews2016underwater, berman2020underwater, akkaynak2019sea} to estimate clean images, aiming to correct global color casts and hazy appearances. Recent approaches adopt learning-based techniques~\cite{li2019underwater, badran2023daut, wang2023domain}, with some further leveraging 3D structural cues for enhanced correction~\cite{boittiaux2024sucre}. For temporal degradations such as caustics and flickering, illumination variations are often modeled as additive~\cite{garcia2002way} or multiplicative~\cite{shihavuddin2012online, gracias2008motion} components and removed via homomorphic filtering. Other works~\cite{sauder2024self, agrafiotis2018underwater} exploit multi-view consistency based on 3D geometry to suppress temporal inconsistencies. 

In underwater 3D reconstruction, the coupling of spatial and temporal degradations presents significant challenges. NeRF-based approaches~\cite{levy2023seathru, tang2024neural, zhang2023beyond, ramazzina2023scatternerf} simulate light propagation volumetrically to model spatial degradations, but often suffer from inefficiency. Recent 3DGS-based methods~\cite{li2025watersplatting, yang2024seasplat} incorporate imaging formation models into the 3DGS framework and improve rendering efficiency, yet still primarily address spatial degradation while neglecting transient lighting effects. In contrast, RecGS~\cite{zhang2024recgs} tackles flickering through recurrent modeling in temporal sequences, but omits spatial degradations. Unlike prior work that targets either spatial or temporal degradations in isolation, our MarineSTD-GS jointly models both within a unified framework, enabling robust and faithful underwater scene reconstruction.

\begin{figure*}[h]
  \centering
  \includegraphics[width=0.9\linewidth]{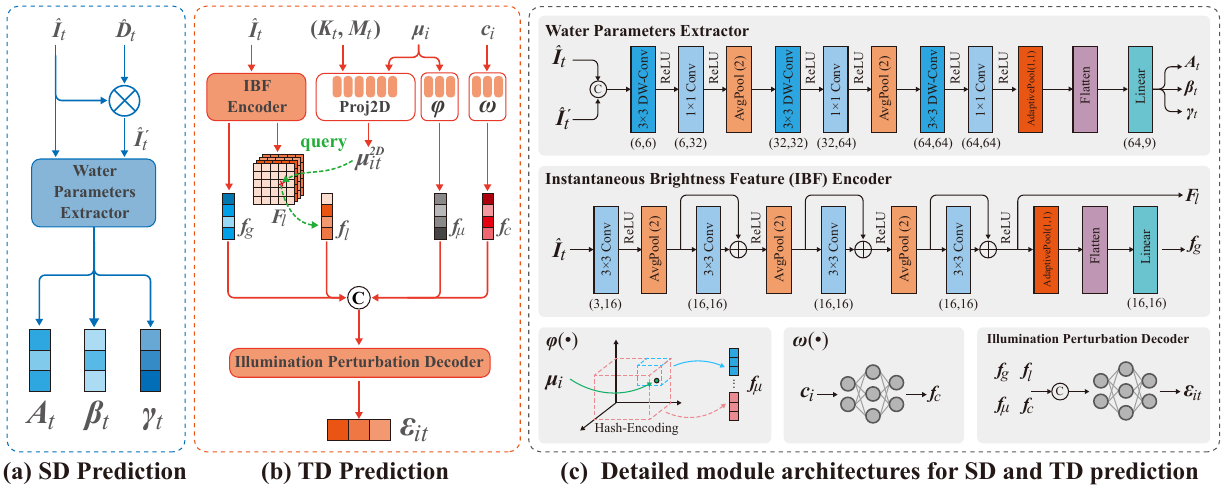}
  \caption{Architectures of (a) SD Prediction branch, (b) TD Prediction branch, and (c) their detailed module architectures.}
  \Description{Architectures of the SD and TD prediction branches and their detailed architectures.}
\label{fig:SDM}
\end{figure*}

\section{Method}
\label{sec:Method}

\subsection{Overview}
\label{sec:Overview}

The pipeline of MarineSTD-GS is illustrated in Fig.~\ref{fig:pipeline}. To enable self-supervised learning of the 3D representation of realistic underwater scenes, we design two paired types of 3D Gaussian primitives: \textbf{Intrinsic Gaussians} and their corresponding \textbf{Degraded Gaussians}. Each pair shares the same geometric attributes (i.e., position, covariance, and opacity), while their color attributes are functionally decoupled. The Intrinsic Gaussians are designed to represent the underlying scene content, and their color attributes aim to capture the realistic appearance of the scene. The Degraded Gaussians, in contrast, are used to render the degraded underwater images corresponding to specific training views at given times, serving as the supervisory signal for learning. The color attributes of the Degraded Gaussians are derived from those of their paired Intrinsic ones through the Spatiotemporal Degradation Modeling (SDM) module, which physically models the spatiotemporal color degradation process.

Below, we detail the SDM module (Section~\ref{sec:SDModeling}), the loss functions (Section~\ref{sec:Loss}), and the optimization strategy (Section~\ref{sec:MSO}).

\subsection{Spatiotemporal Degradation Modeling}
\label{sec:SDModeling}

As illustrated in Fig.~\ref{fig:pipeline}, the SDM module comprises a physically grounded Color Degradation Equation that models how scene colors degrade under varying spatiotemporal conditions, and two dedicated prediction branches that respectively estimate distinct types of degradation parameters required by the equation.

\subsubsection{Color Degradation Equation}
\label{sec:CDE}
For the $i$-th pair of Gaussians, we model the relationship between the color $\bm{c}_{i}$ of the Intrinsic Gaussian and the color $\bm{c}_{it}$ of its corresponding Degraded Gaussian at time $t$ as follows:
\begin{equation}
\label{eq:cde}
\bm{c}_{it}=\bm{c}_{i} \underbrace{
e^{-\bm{\beta}_{t} \cdot  r_{it}} + \bm{A}_{t} (1 - e^{-\bm{\gamma}_{t}  \cdot  r_{it}})
}_{\text{SD term}} + 
\underbrace{
\epsilon_{it}
}_{\text{TD term}}
.
\end{equation}
Equation~\ref{eq:cde} consists of two components: the SD term and the TD term. The SD term models the effects of spatial degradation (SD), such as attenuation and backscattering, while the TD term captures the influence of temporal degradation (TD) caused by dynamic lighting phenomena like caustics and flickering. Specifically, the SD term is inspired by the underwater image formation model proposed by Akkaynak~\textit{et al.}. It incorporates three water parameters that characterize the optical properties of the medium: the attenuation coefficient $\bm{\beta}_{t}$, the backscatter coefficient $\bm{\gamma}_{t}$, and the ambient underwater light $\bm{A}_{t}$. These water parameters are shared across all Gaussian pairs at time $t$ and are predicted by the SD Prediction branch. Here, $r_{it}$ denotes the distance between the $t$-th viewpoint and the $i$-th Gaussian. In contrast, the TD term models local temporal degradations as additive illumination variations, using a Gaussian-specific perturbation term $\epsilon_{it}$, which is predicted by the TD Prediction branch.

\subsubsection{Spatial Degradation Prediction}
\label{sec:SDP}
As shown in Fig.\ref{fig:SDM}(a), the SD Prediction branch directly predicts the water parameters from the degraded image $\hat{\bm{I}}_t$. To better exploit distant regions where water effects are more significant for accurate prediction, we apply a depth-aware enhancement using the pseudo-depth map $\hat{\bm{D}}_t$ generated by Depth-Anything-V2\cite{yang2024depthv2}, producing an enhanced image  $\hat{\bm{I}}_t' = \hat{\bm{D}}_t \otimes \hat{\bm{I}}_t$. We then concatenate $\hat{\bm{I}}_t'$ and $\hat{\bm{I}}_t$ along the channel dimension and feed them into a Water Parameters Extractor (WPE) to estimate the final $\bm{A}_{t}$, $\bm{\beta}_{t}$, and $\bm{\gamma}_{t}$. The detailed structure of WPE is shown in Fig.~\ref{fig:SDM}(c).

\subsubsection{Temporal Degradation Prediction}
\label{sec:TDP}
As shown in Fig.~\ref{fig:SDM}(b), to predict the local instantaneous brightness variation of the $i$-th Gaussian at time $t$, we employ an Instantaneous Brightness Feature (IBF) Encoder to extract a low-resolution feature map $\bm{F}_l$ and a global feature vector $\bm{f}_g$ from the degraded image $\hat{\bm{I}}_t$. While $\bm{F}_l$ captures spatially uneven brightness patterns caused by flickering or caustics, $\bm{f}_g$ encodes global illumination. Then, the 2D projection coordinate $\mu_{it}^{2D}$ of the $i$-th Gaussian under the $t$-th view is computed via $\mu_{it}^{2D} = \text{Proj2D}(\bm{K}_t, \bm{M}_t, \mu_i)$, based on the camera intrinsics $\bm{K}_t$ and extrinsics $\bm{M}_t$. Using this coordinate, the corresponding local brightness feature vector $\bm{f}_l$ is retrieved from $\bm{F}_l$ via bilinear interpolation. Meanwhile, the intrinsic color $\bm{c}_i$ and 3D position $\bm{\mu}_i$ of the $i$-th Gaussian are encoded into feature vectors using two learnable encoders, $\phi(\cdot)$ and $\omega(\cdot)$, respectively. Finally, all features are concatenated and passed to an Illumination Perturbation Decoder to estimate the transient illumination perturbation $\epsilon_{it}$ of the $i$-th Gaussian at time $t$. Architectural details of the IBF Encoder, $\phi(\cdot)$, $\omega(\cdot)$, and the Decoder are shown in Fig.~\ref{fig:SDM}(c).

\subsection{Loss Function}
\label{sec:Loss}

\subsubsection{Photometric Loss}
In our self-supervised setup, we adopt a photometric loss $L_{\text{photo}}$ from standard 3DGS~\cite{kerbl20233d} to align the rendered image with the degraded input:
\begin{equation}
L_{\text{photo}} = \lambda_1 \cdot \| \hat{\bm{I}_t} - \bm{I}_t \|_1 + \lambda_2 \cdot \text{D-SSIM}(\hat{\bm{I}}_t, \bm{I}_t),
\end{equation}
where $L_{\text{photo}}$ combines L1 and D-SSIM terms, with weights $\lambda_1$ and $\lambda_2$. $\hat{\bm{I}}_t$ and $\bm{I}_t$ denote the rendered and degraded input images at time $t$, respectively.


\begin{figure*}[h]
\centering
\includegraphics[width=\linewidth]{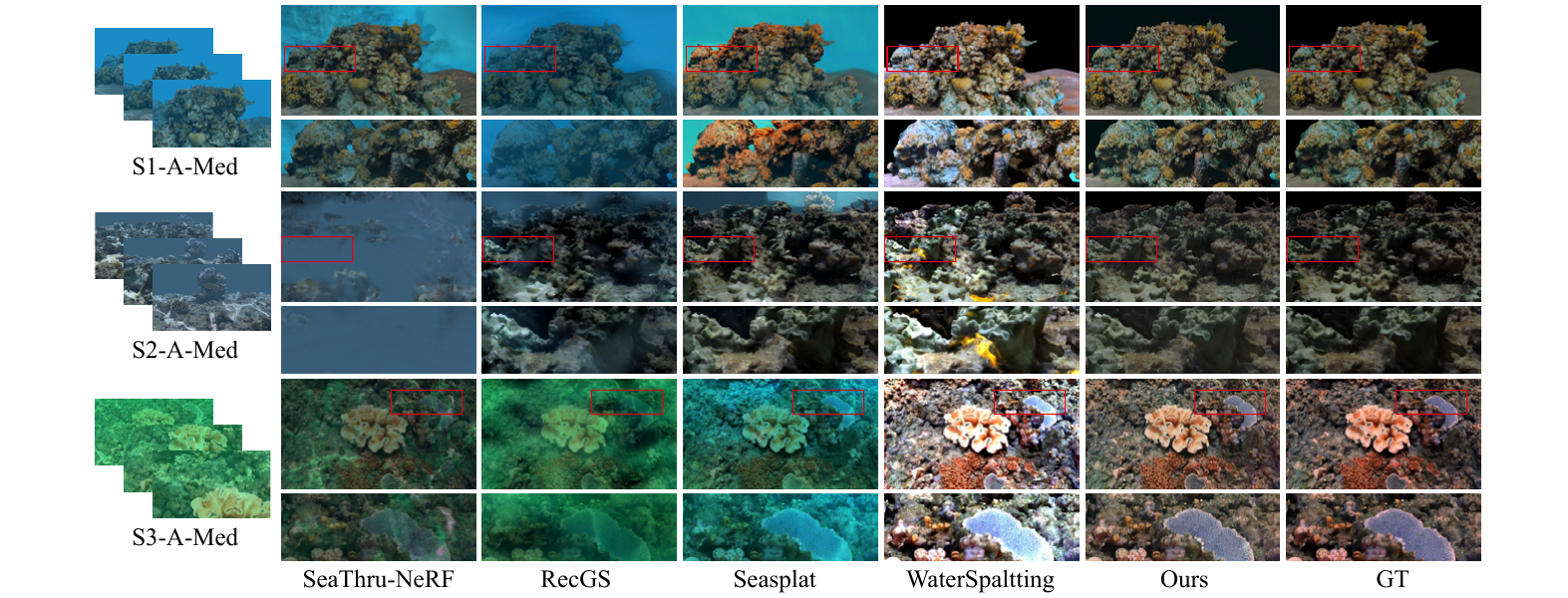}
\caption{
Qualitative novel view synthesis results on simulated scenes.
Our method removes global color distortions and local overexposures, producing appearances closest to the ground truth.
}
\Description{Comparison of novel view synthesis results from different methods on synthetic underwater scenes. Our method best restores scene colors and suppresses caustic-related artifacts, closely matching the ground truth.}
\label{fig:nvs_syn}
\end{figure*}

\begin{table*}
\caption{
Quantitative evaluation of novel view synthesis on simulated scenes, with results reported separately for each scene category. 
The best results are highlighted in \textbf{bold}, and the second-best are \underline{underlined}. 
Efficiency metrics are also reported.
}

  \label{tab:nvs_syn}
  
\begin{tabular}{c|c|ccc|ccc|ccc|c|c}
    \toprule
\multirow{2}{*}{Methods} & \multirow{2}{*}{Paper/Year} & \multicolumn{3}{c|}{S1 Scene}                     & \multicolumn{3}{c|}{S2 Scene}                     & \multicolumn{3}{c|}{S3 Scene}                     & \multirow{2}{*}{FPS} & \multirow{2}{*}{\begin{tabular}[c]{@{}c@{}}Avg.\\ Time\end{tabular}} \\ 
\cmidrule(){3-11}

                         &                             & PSNR            & SSIM           & LPIPS          & PSNR            & SSIM           & LPIPS          & PSNR            & SSIM           & LPIPS          &                      &                                                                      \\ 
    \midrule
                         
3DGS                     & SIGGRAPH/2023               & 17.656          & 0.792          & 0.167          & 15.218          & 0.695          & 0.163          & 13.804          & 0.692          & 0.578          & 125.43               & 0.40h                                                                \\
gsplat                   & JMLR/2024                  & 17.871          & 0.782          & 0.179          & 15.257          & 0.685          & 0.183          & 13.741          & 0.673          & 0.599          & 107.24               & 0.16h                                                                \\
GS-W                     & ECCV/2024                   & 18.010          & 0.778          & 0.198          & 17.001          & 0.687          & 0.220          & 14.340          & 0.666          & 0.570          & 67.68                & 1.18h                                                                \\
Splatfacto-W              & Arxiv/2024                  & 16.933          & 0.772          & 0.193          & 14.176          & 0.663          & 0.195          & 14.132          & 0.695          & 0.581          & 91.42                & 0.16h                                                                \\
SeaThru-NeRF             & CVPR/2023                   & 23.995          & 0.885          & 0.105          & 15.348          & 0.529          & 0.420          & 15.934          & 0.711          & 0.358          & 0.23                 & 13.5h                                                                \\
Seasplat                 & Arxiv/2024                  & {\ul {24.064}} & {\ul {0.908}} & 0.097          & 18.292          & {\ul {0.813}} & {\ul {0.124}} & {\ul {17.379}} & {\ul {0.830}} & 0.294          & 37.71                & 0.52h                                                                \\
RecGS                    & ICRA/2025                   & 18.576          & 0.777          & 0.159          & {\ul {18.551}} & 0.728          & 0.164          & 14.230          & 0.685          & 0.549          & 392.16               & 0.33h                                                                \\
WaterSpaltting           & 3DV/2025                    & 18.739          & 0.882          & {\ul {0.075}} & 14.279          & 0.767          & 0.137          & 14.275          & 0.809          & {\ul {0.152}} & 84.53                & 0.12h                                                                \\ 
    \midrule

\textbf{Ours}         & \textbf{ACM MM/2025}                     & \textbf{27.719} & \textbf{0.934} & \textbf{0.059} & \textbf{25.853} & \textbf{0.863} & \textbf{0.085} & \textbf{26.355} & \textbf{0.918} & \textbf{0.097} & 104.91               & 0.38h                                                                \\ 
    \bottomrule

\end{tabular}

\end{table*}

\subsubsection{Depth-Guided Geometry Loss}

Due to spatiotemporal degradation breaking multi-view consistency, photometric loss alone becomes insufficient for accurate geometry learning. Recent monocular depth estimators, such as Depth-Anything-V2~\cite{yang2024depthv2}, remain robust even under severe caustic-induced degradations. Inspired by this, we introduce a Depth-Guided Geometry Loss ($L_{\text{dgg}}$) that leverages pseudo-depth priors for improved reconstruction, consisting of a coarse depth term and an adaptive edge-aware depth smoothness term.

\textbf{Coarse Depth Supervision Term.} Since pseudo-depth maps reflect normalized disparities rather than absolute depths, direct supervision using L1/L2 loss is unsuitable. Instead, we adopt the Pearson correlation coefficient~\cite{xiong2023sparsegs}, which is translation- and scale-invariant, to measure similarity between the pseudo-depth $\bm{D}_t$ and the rendered depth $\hat{\bm{D}}t$:
\begin{equation}
L_{\text{cds}} = 1 - \text{Pearson}(\bm{D}_t, \hat{\bm{D}}_t).
\end{equation}

\textbf{Adaptive Edge-aware Depth Smoothness Term.} While $L_{\text{cds}}$ provides coarse global geometry supervision, local geometric consistency still requires further regularization. To this end, we introduce an adaptive edge-aware depth smoothness term $L_{\text{ads}}$. It computes spatial gradients of the rendered depth $\bm{D}_t$, weighted by adaptive factors $\bm{w}_x$ and $\bm{w}_y$ derived from the pseudo-depth $\hat{\bm{D}}_t$ and input image $\bm{I}_t$:
\begin{equation}
L_{\text{ads}} = \frac{1}{N} \sum_{x,y} \left( |\nabla_x \bm{D}_t \cdot \bm{w}_x| + |\nabla_y \bm{D}_t \cdot \bm{w}_y| \right),
\end{equation}
where $\nabla_x \bm{D}_t$, $\bm{w}_x$, etc., are evaluated at each pixel $(x, y)$ and the weights are defined as:
\begin{equation}
\bm{w}_{\{x,y\}} = (1 - \hat{\bm{D}}_t) \cdot e^{-|\nabla_{\{x,y\}} \hat{\bm{D}}_t|} + \hat{\bm{D}}_t \cdot e^{-|\nabla_{\{x,y\}} \hat{\bm{I}}_t|},
\end{equation}
where $\nabla_{{x,y}}$ denotes spatial gradients along $x$ or $y$ at pixel $(x, y)$, and $\hat{\bm{D}}_t \in [0,1]$ assigns higher values to distant regions, thereby relying more on RGB-based edge cues in regions where depth is less reliable.

The final Depth-Guided Geometry Loss is defined as $L_{\text{dgg}} = \lambda_{\text{cds}} L_{\text{cds}} + \lambda_{\text{ads}} L_{\text{ads}}$, where $\lambda_{\text{cds}}$ and $\lambda_{\text{ads}}$ are the corresponding weights.

\subsubsection{Total Loss}

In addition to the photometric loss and the Depth-Guided Geometry Loss, the final training objective incorporates a regularization term $L_{\epsilon\text{-reg}}$ on the transient illumination perturbations $\epsilon_{it}$. Specifically, we apply an $\ell_2$ penalty to all predicted $\epsilon_{it}$ values to prevent the SDM module from overfitting to color reconstruction errors by over-relying on the transient term. This regularization helps avoid trivial solutions, stabilizes the disentanglement between intrinsic color and transient lighting, and mitigates cross-talk between the SD and TD branches by encouraging each to learn its respective degradation component more accurately. The total loss is defined as:
\begin{equation}
L_{\text{total}} = L_{\text{photo}} + L_{\text{dgg}} + \lambda_{\epsilon} \cdot L_{\epsilon\text{-reg}},
\end{equation}
where $L_{\epsilon\text{-reg}} = \sum_{i,t} || \epsilon_{it} ||_2^2$, and $\lambda_{\epsilon}$ controls its strength.

\subsection{Multi-Stage Optimization}
\label{sec:MSO}

In practice, we observe that transient illumination perturbations predicted by the TD branch may interfere with backscatter estimation from the SD branch, leading to optimization conflicts within the SDM module. To stabilize training, we adopt a multi-stage optimization strategy consisting of three phases:

\textbf{Stage I: Warm-up.} The TD branch participates in degraded color computation but remains frozen (no gradient updates), allowing the SD branch to independently learn reasonable water parameters without interference.

\textbf{Stage II: Iterative training.} The TD and SD branches are alternately updated in an interleaved manner—at each iteration, one branch is optimized while the other remains fixed, providing a stable degradation signal. This mitigates interference and facilitates decoupled learning of the two degradation types.

\textbf{Stage III: Joint training.} Both branches are jointly optimized to refine parameter learning and enhance disentanglement, improving reconstruction quality and realism.

After training, the Intrinsic Gaussians effectively capture the underlying scene and can be directly used for novel view synthesis as if observed without underwater degradations such as attenuation, backscatter, caustics, or flickering.

\begin{figure}[h]
\centering
\includegraphics[width=\linewidth]{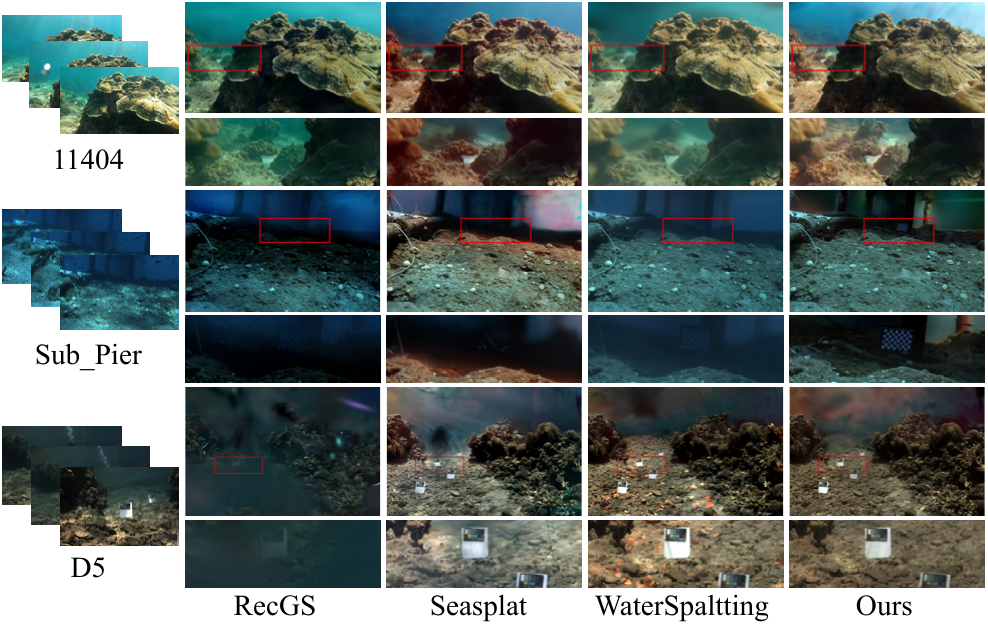}
\caption{
Qualitative novel view synthesis results on real-world scenes.
Our method corrects color casts, removes haze effects, and suppresses caustic-induced illumination artifacts, yielding a consistent appearance across views.
}
\Description{Side-by-side comparison of novel view synthesis results from different methods on real-world underwater scenes. Our method shows clearer colors, reduced haze, and fewer lighting artifacts.}
\label{fig:nvs_real}
\end{figure}

\begin{figure}[h]
\centering
\includegraphics[width=\linewidth]{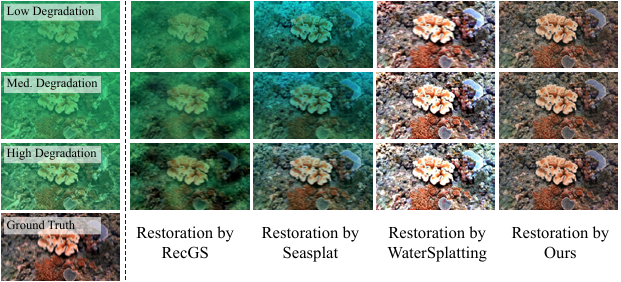}
\caption{
Qualitative comparisons under different spatial degradation levels on simulated scenes. Our method consistently preserves scene appearance and color fidelity, even under severe degradation, closely matching the ground truth.
}
\Description{Comparison of novel view synthesis results under low, medium, and high spatial degradation levels on simulated scenes. Our method shows clearer scene structure and more accurate colors, closely resembling the ground truth, while other methods exhibit color shifts and blurring as degradation increases.}
\label{fig:Rebust}
\end{figure}

\begin{figure}[h]
\centering
\includegraphics[width=\linewidth]{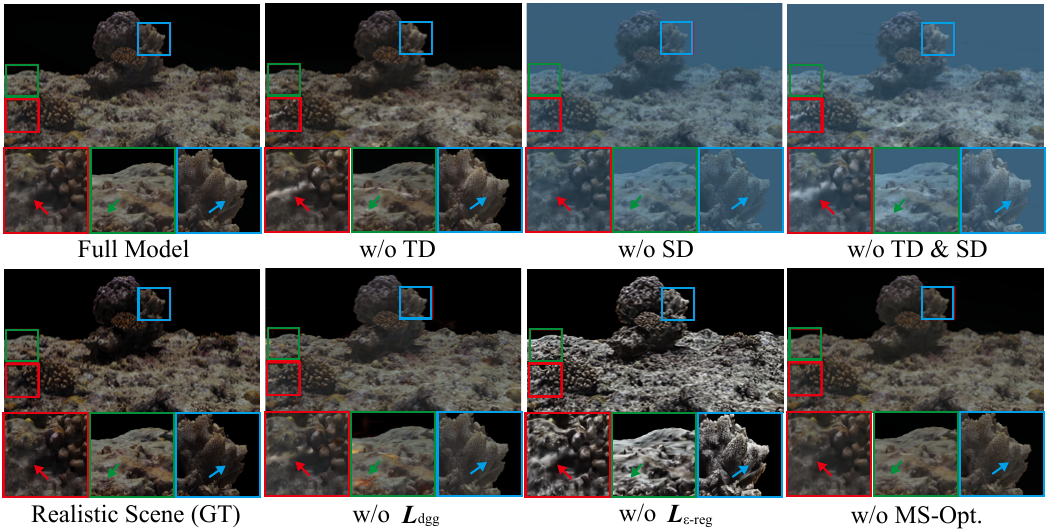}
\caption{Visual comparisons of novel view synthesis under different ablation settings.}
\Description{Comparison of synthesized views with different ablation settings.}
\label{fig:Ablation_rgb}
\end{figure}

\begin{figure}[h]
\centering
\includegraphics[width=\linewidth]{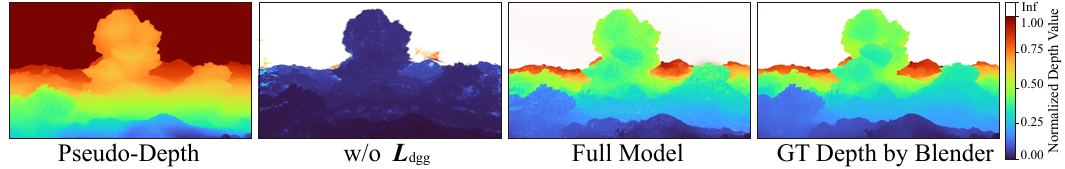}
\caption{Effectiveness of depth-guided geometry supervision with pseudo-depth. }
\Description{Comparison of geometry prediction with and without pseudo-depth guidance. Results show improved accuracy and robustness when pseudo-depth is used for supervision.}
\label{fig:Ablation_depth}
\end{figure}

\section{Experiments}
\label{sec:Experiments}

\subsection{Experimental Setup}
\label{sec:setup}

\textbf{Simulated Dataset.}
Since obtaining real underwater scenes without degradation is nearly impossible, we construct three categories of synthetic underwater scenes using Blender, each incorporating spatial and temporal degradation factors for comprehensive evaluation. The three scene categories are designed as follows:  
\textit{S1} contains detailed textures and diverse colors;
\textit{S2} simulates large-scale open environments with contrast loss in distant regions;
\textit{S3} features dominant greenish color distortions.
For each category, we generate five variants: three with different caustic patterns (Pattern A/B/C) under medium spatial degradation, and two with fixed Pattern A but different spatial degradation levels (low and high). This design enables systematic evaluation under varying temporal degradation patterns and spatial degradation levels. Each scene contains 120 images at a resolution of 540$\times$960.

\textbf{Real-world Dataset.}
We evaluate our method on a variety of real underwater scenes, including four representative scenes from the BVICoral~\cite{anantrasirichai_2024_11093417} dataset and four from the Flseas\_VI~\cite{randall2023flsea} dataset. These scenes exhibit strong caustics, flickering, and significant color degradation. Additionally, we include four scenes from the SeaThru-NeRF dataset~\cite{levy2023seathru} and two scenes (D3 and D5) from the SeaThru dataset~\cite{akkaynak2019sea}. More details on the simulated and real-world datasets are provided in Appendix~\ref{app:dataset}.

\textbf{Comparison Methods.}
We compare our method with eight state-of-the-art baselines, including four general 3D reconstruction methods (3DGS~\cite{kerbl20233d}, gsplat~\cite{ye2025gsplat}, GS-W~\cite{zhang2024gaussian}, and Splatfacto-W~\cite{xu2024splatfacto}), and four underwater-specific approaches (SeaThru-NeRF~\cite{levy2023seathru}, WaterSplatting~\cite{li2025watersplatting}, SeaSplat~\cite{yang2024seasplat}, and RecGS~\cite{zhang2024recgs}).

\textbf{Evaluation Metrics.}
We evaluate novel view synthesis on realistic scenes restored from underwater degradation. For simulated data, we compute PSNR, SSIM, and LPIPS between rendered and ground-truth images. For real-world scenes with color charts, we assess color fidelity using CIEDE2000 ($\Delta E_{00}$)\cite{gaurav2005ciede2000} and average angular error $\bar{\psi}$ (in degrees)\cite{boittiaux2024sucre}. We also report average training time (Avg. Time) and rendering speed (FPS) at 540$\times$960 resolution to reflect efficiency.

\textbf{Implementation Details.}
Our method builds on gsplat~\cite{ye2025gsplat}. 3D Gaussians are initialized from sparse point clouds reconstructed by COLMAP~\cite{schonberger2016structure}, with intrinsic colors from corresponding point colors. To reflect the Lambertian nature of underwater scenes~\cite{murez2015photometric}, we set the SH degree to zero. We adopt the multi-stage training strategy (Sec.~\ref{sec:MSO}) for 30{,}000 steps in total (10{,}000 per stage). Loss weights are set as: $\lambda_1 = 0.8$, $\lambda_2 = 0.2$ for $L_{\text{photo}}$; $\lambda_{\text{cds}} = 0.1$, $\lambda_{\text{ads}} = 0.01$ for $L_{\text{dgg}}$; and $\lambda_{\epsilon} = 100$ for $L_{\epsilon\text{-reg}}$. All experiments run on a single NVIDIA RTX 3090 GPU. Additional implementation details are provided in Appendix~\ref{app:implementation}.

\begin{table}
\caption{Color Correction Results on Real Scenes.}
\setlength{\tabcolsep}{1.2pt}
\label{tab:nvs_real}
\begin{tabular}{c|cc|cc|cc|cc}

\toprule

\multirow{2}{*}{Methods} & \multicolumn{2}{c|}{Curasao}     & \multicolumn{2}{c|}{D3}          & \multicolumn{2}{c|}{D5}          & \multicolumn{2}{c}{Avg.}         \\ 
\cmidrule(){2-3} \cmidrule(){4-5} \cmidrule(){6-7} \cmidrule(){8-9}

                         & $\Delta E_{00}$ & $\bar{\psi}$   & $\Delta E_{00}$ & $\bar{\psi}$   & $\Delta E_{00}$ & $\bar{\psi}$   & $\Delta E_{00}$ & $\bar{\psi}$   \\ 
\midrule
                         
3DGS                     & 25.63           & 26.35          & 20.18           & 22.65          & 32.38           & 26.62          & 26.06           & 25.21          \\
gsplat                   & 23.24           & 26.20          & 20.34           & {\ul {22.45}} & 30.93           & 26.52          & {\ul {24.84}}   & 25.06          \\
GS-W                     & 25.76           & 26.45          & 27.82           & 26.66          & 35.17           & 27.64          & 29.58           & 26.92          \\
Splatfacto-W              & 28.31           & 26.92          & 22.12           & 23.29          & 31.66           & 27.45          & 27.36           & 25.89          \\
SeaThru-NeRF             & {\ul {20.97}}  & \textbf{23.29} & 21.13           & 24.92          & 34.44           & 26.92          & 25.51           & 25.04          \\
Seasplat                 & 24.33           & 23.59          & {\ul {19.86}}  & 22.80          & 37.04           & 27.86          & 27.08           & 24.75          \\
RecGS                    & 39.53           & 29.00          & 20.15           & 22.55          & 24.09           & 23.66          & 27.92           & 25.07          \\
WaterSpaltting           & 37.16           & 26.53          & 24.30           & 23.05          & \textbf{20.45}  & {\ul {21.83}} & 27.30           & {\ul {23.80}} \\
\midrule

\textbf{Ours}            & \textbf{19.98}  & {\ul {23.40}} & \textbf{19.64}  & \textbf{22.27} & {\ul {22.23}}   & \textbf{20.95} & \textbf{20.62}  & \textbf{22.21} \\

\bottomrule

\end{tabular}
\end{table}

\begin{table}
\caption{Quantitative evaluation of robustness under varying spatial degradation levels. }
\setlength{\tabcolsep}{3pt}
\label{tab:rebust}
\begin{tabular}{c|cc|cc|cc}
\toprule

\multirow{2}{*}{Methods} & \multicolumn{2}{c|}{Low} & \multicolumn{2}{c|}{Medium} & \multicolumn{2}{c}{High} \\

\cmidrule(){2-7}
                         & PSNR              & SSIM             & PSNR                & SSIM              & PSNR              & SSIM             \\
\midrule                         
3DGS                     & 17.417            & 0.808            & 15.623              & 0.729             & 13.988            & 0.638            \\
gsplat                   & 17.470            & 0.793            & 15.811              & 0.721             & 14.073            & 0.629            \\
GS-W                     & 17.862            & 0.789            & 16.131              & 0.716             & 14.383            & 0.634            \\
Splatfacto-W              & 16.065            & 0.771            & 15.214              & 0.715             & 13.970            & 0.637            \\
SeaThru-NeRF             & 18.479            & 0.705            & 18.152              & 0.687             & {\ul {18.289}}   & 0.700            \\
Seasplat                 & {\ul {21.535}}   & {\ul {0.895}}   & {\ul {19.430}}     & {\ul {0.839}}    & 17.242            & 0.780            \\
RecGS                    & 19.119            & 0.789            & 17.391              & 0.731             & 15.245            & 0.645            \\
WaterSpaltting           & 14.367            & 0.781            & 15.379              & 0.812             & 16.800            & {\ul {0.835}}   \\
\midrule
\textbf{Ours}         & \textbf{27.520}   & \textbf{0.927}   & \textbf{27.502}     & \textbf{0.913}    & \textbf{25.396}   & \textbf{0.875}   \\ 
\bottomrule

\end{tabular}
\end{table}

\begin{table}
\caption{Quantitative evaluation of robustness under different temporal degradation Patterns. }
\setlength{\tabcolsep}{3pt}
\label{tab:rebust_TD}

\begin{tabular}{c|cc|cc|cc}
\toprule
\multirow{2}{*}{Methods} & \multicolumn{2}{c|}{Pattern A}   & \multicolumn{2}{c|}{Pattern   B} & \multicolumn{2}{c}{Pattern   C}  \\ \cmidrule(){2-7}
                         & PSNR            & SSIM           & PSNR            & SSIM           & PSNR            & SSIM           \\ \midrule 
3DGS                     & 15.623          & 0.729          & 15.122          & 0.728          & 15.645          & 0.728          \\
gsplat                   & 15.811          & 0.721          & 15.068          & 0.707          & 15.693          & 0.716          \\
GS-W                     & 16.131          & 0.716          & 16.770          & 0.709          & 17.105          & 0.703          \\
Splatfacto-W             & 15.214          & 0.715          & 14.931          & 0.713          & 15.222          & 0.713          \\
SeaThru-NeRF             & 18.152          & 0.687          & 18.911          & 0.744          & 18.297          & 0.706          \\
Seasplat                 & {\ul {19.430}}    & {\ul {0.839}}    & {\ul {20.511}}    & {\ul {0.864}}    & {\ul {20.843}}    & {\ul {0.873}}  \\
RecGS                    & 17.391          & 0.731          & 16.541          & 0.751          & 17.300          & 0.734          \\
WaterSpaltting           & 15.379          & 0.812          & 16.289          & 0.839          & 15.988          & 0.830          \\ \midrule 
\textbf{Ours}            & \textbf{27.502} & \textbf{0.913} & \textbf{25.757} & \textbf{0.900} & \textbf{27.037} & \textbf{0.909} \\ \bottomrule
\end{tabular}

\end{table}

\begin{table}
\caption{Quantitative evaluation of ablation experiments on simulated scenes under Caustic Pattern A. }
\setlength{\tabcolsep}{12pt}
\centering
\label{tab:Ablation}
\begin{tabular}{l|ccc}
\toprule
Methods                        & PSNR             & SSIM             & LPIPS            \\
\midrule
w/o SD                         & 16.804           & 0.730            & 0.310            \\
w/o TD                         & 24.440           & 0.874            & 0.111            \\
w/o TD \& SD                   & 15.876           & 0.718            & 0.315            \\
\midrule
w/o $L_{\text{dgg}}$           & \ul{25.791}      & \ul{0.893}       & \ul{0.086}       \\
w/o $L_{\epsilon\text{-reg}}$ & 19.388           & 0.774            & 0.194            \\
w/o MS-Opt.                    & 25.705           & 0.887            & 0.094            \\
\midrule
\textbf{Full Model}                     & \textbf{26.806}  & \textbf{0.905}   & \textbf{0.080}   \\
\bottomrule
\end{tabular}

\end{table}

\subsection{Experimental Results}

\textbf{Quantitative Results on Simulated Scenes.} Table~\ref{tab:nvs_syn} reports the quantitative results of novel view synthesis on the simulated dataset. Our method consistently achieves the best performance across all metrics and scene categories. Notably, on the S1 scenes, it outperforms the second-best method (Seasplat) by 3.655 dB in PSNR. For S3, our advantage increases to 8.976 dB, indicating strong robustness to severe color distortions.

\textbf{Quantitative Results on Real-world Scenes.} Table~\ref{tab:nvs_real} presents quantitative color correction results on real-world scenes. Our method leads on both evaluation metrics, reducing $\Delta E_{00}$ by 4.22 compared to gsplat, and lowering the angular error $\bar{\psi}$ by 1.59 compared to WaterSplatting, demonstrating its effectiveness in real-world color fidelity recovery.

\textbf{Qualitative Comparisons.} Fig.~\ref{fig:nvs_syn} and Fig.~\ref{fig:nvs_real} show qualitative comparisons. RecGS, which ignores spatial degradations, fails to remove the color cast and haze caused by attenuation and backscattering. SeaThru-NeRF and SeaSplat introduce strong color artifacts, while WaterSplatting struggles to suppress local overexposure under caustics. In contrast, our method corrects color distortions, enhances visibility, and suppresses caustic-induced illumination artifacts across both simulated and real-world scenes. For example, in the Sub\_Pier and D5 scenes, our method restores distant structures and produces consistent color and lighting.

\textbf{Robustness Under Different Spatial Degradation Levels and Temporal Degradation Patterns.}
Table~\ref{tab:rebust} compares the robustness of all methods under different spatial degradation levels. Our method consistently outperforms others, maintaining high fidelity even under severe degradation, with at least a 5.985 dB PSNR margin over the second-best across all levels. Fig.~\ref{fig:Rebust} further illustrates that our reconstructed scenes best preserve color fidelity and appearance relative to the ground truth, regardless of the spatial degradation level. Table~\ref{tab:rebust_TD} further evaluates robustness under three temporal degradation patterns (A/B/C). Our method consistently achieves the best performance in both PSNR and SSIM, demonstrating strong adaptability to dynamic degradations.

\subsection{Ablation Study}

All ablation experiments are conducted on all three simulated scenes under Caustic Pattern A, covering all spatial degradation levels.

\textbf{Effectiveness of SDM Components.}  
We assess the two core branches of the SDM module: TD Prediction and SD Prediction. As shown in Table~\ref{tab:Ablation}, removing either branch—or both—leads to a PSNR drop of at least 2.366 dB. Fig.~\ref{fig:Ablation_rgb} shows that removing the TD branch results in flickering-induced local highlights, while removing the SD branch fails to correct water degradation. These results confirm the necessity of both branches in modeling and disentangling their respective degradation types.

\textbf{Impact of Losses and Training Strategy.}  
We evaluate the impact of the transient regularization loss $L_{\epsilon\text{-reg}}$, the depth-guided geometry loss $L_{\text{dgg}}$, and the multi-stage optimization strategy (MS-Opt.). As shown in Table~\ref{tab:Ablation}, all three contribute positively to overall performance. Fig.~\ref{fig:Ablation_rgb} shows that without $L_{\epsilon\text{-reg}}$, the model overfits to the transient term, causing the Intrinsic Gaussians to degenerate into overly monotonic colors; without MS-Opt., the restored appearance becomes less detailed and less stable. Fig.~\ref{fig:Ablation_depth} further demonstrates the benefit of $L_{\text{dgg}}$: although the pseudo-depth may deviate from the ground truth, it still effectively guides geometry reconstruction toward structurally accurate results, indicating the robustness of our approach to noise in pseudo-depth priors.

\begin{figure}[h]
\centering
\includegraphics[width=\linewidth]{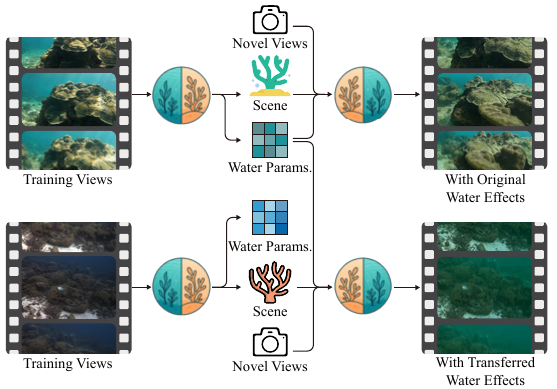}
\caption{
Applications of disentangled realistic scene and water representations. 
Top: Novel view synthesis with original water effects. 
Bottom: Water effect transfer for novel view synthesis under different water conditions.
}
\Description{
The top row shows novel views rendered using the original scene’s water effects. The bottom row demonstrates transferring estimated water effects from one scene to another, generating novel underwater views with new media conditions.
}
\label{fig:transfer}
\end{figure}

\subsection{More Applications}
\label{sec:Applications}
Thanks to our disentanglement of clean 3D scene representations and per-frame spatial degradation-aware water parameters from underwater video sequences, our method supports not only clean novel view synthesis but also two immersive underwater rendering modes with realistic attenuation and backscattering effects (Fig.~\ref{fig:transfer}). First, we render stable underwater views by applying averaged water parameters from a sequence back to the same scene. Second, we support cross-scene water effect transfer, where a source scene is rendered using water parameters extracted from a different reference video, enabling underwater view synthesis under varied media conditions. These capabilities support environment simulation, data generation, and synthetic dataset construction with known geometry.

\section{Conclusion}
We propose MarineSTD-GS, a 3D Gaussian framework that explicitly models spatiotemporal degradation in underwater videos. With a dual-Gaussian design and a physically grounded SDM module, it jointly learn realistic clean scenes and associated degradation factors via self-supervision learning.  The proposed Depth-Guided Geometry Loss improves geometric accuracy, while the Multi-Stage Optimization strategy stabilizes training and enhances texture fidelity. Experiments on both simulated and real-world underwater scenes show that MarineSTD-GS achieves state-of-the-art results in color fidelity, visibility, and transient artifact suppression. Our disentangled modeling also supports novel view synthesis with controllable water effects, highlighting its potential for simulation, virtual reality, and data generation in underwater environments.


\begin{acks}
This work was supported by the National Natural Science Foundation of China under Grants 62405014, 62325101, and 62031001.
\end{acks}

\bibliographystyle{ACM-Reference-Format}
\balance
\bibliography{sample-base}

\clearpage
\appendix

\section{Dataset Details}
\label{app:dataset}

\subsection{Simulated Dataset}
\label{app:dataset_simulated}

\begin{figure*}[t]
  \centering
  \includegraphics[width=0.8\textwidth]{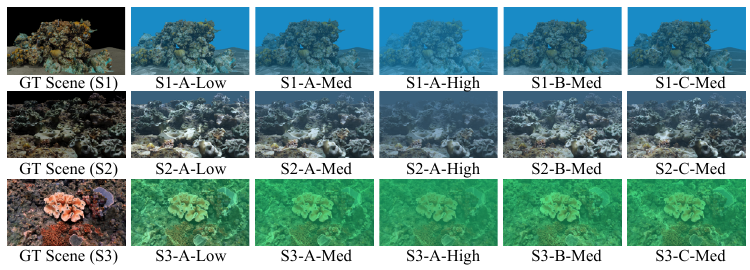}
  \caption{The naming and configuration of our simulated dataset.}
  \Description{The figure illustrates the naming rules and degradation configurations used in the simulated underwater dataset.}
  \label{fig:app_dataset_syn}
\end{figure*}

Since real degradation-free underwater captures are difficult to obtain, we build a simulated benchmark for controlled evaluation, as illustrated in Fig.~\ref{fig:app_dataset_syn}. The benchmark covers both spatial and temporal degradations across three representative scene categories: \textbf{S1}, with rich textures and color variation; \textbf{S2}, a large-scale environment with strong distance-dependent contrast decay; and \textbf{S3}, dominated by greenish color distortions.

We source three underwater 3D models and refine their textures to represent in-air ground-truth appearance. The models are imported into Blender~\cite{blender}, where we design forward-facing camera trajectories to mimic divers or autonomous underwater vehicles. To simulate temporal degradations, we use surface-emission planes driven by video clips of synthetic caustic patterns. These animated textures are projected onto the scene to modulate light emission, producing time-varying caustics and flickering. Blender then renders temporally degraded images, clean ground-truth images, depth maps, and camera intrinsics/poses converted to COLMAP~\cite{schonberger2016structure} format.

For spatial degradations, we apply the revised underwater image formation model of Akkaynak et al.~\cite{akkaynak2018revised} to the temporally degraded renderings using the corresponding depth maps and predefined water parameters. For each scene category, we generate five variants: three caustic patterns (A/B/C) under medium spatial degradation, plus low- and high-degradation variants with Pattern A. Each scene contains 120 images at a resolution of 540$\times$960. Detailed parameters are listed in Table~\ref{tab:app_dataset_syn}.

\begin{table*}[t]
\caption{The configuration of the simulated underwater dataset.}
\label{tab:app_dataset_syn}
\setlength{\tabcolsep}{10pt}
\begin{tabular}{c|c|c|c|ccc}
\toprule
Scene Name & GT Scene            & \begin{tabular}[c]{@{}c@{}}Caustic\\ Pattern\end{tabular} & \begin{tabular}[c]{@{}c@{}}Level of\\ Degradation\end{tabular} & \begin{tabular}[c]{@{}c@{}}Ambient\\ Light\end{tabular} & \begin{tabular}[c]{@{}c@{}}Attenuation\\ Coefficient\end{tabular} & \begin{tabular}[c]{@{}c@{}}Backscatter\\ Coefficient\end{tabular} \\
\midrule
S1-A-Low   & \multirow{5}{*}{S1} & A               & Low                  & \multirow{5}{*}{[0.10, 0.55, 0.78]} & [1.65, 1.45, 1.25]      & [1.33, 1.25, 1.20]      \\
S1-A-Med   &                     & A               & Medium               &                                     & [3.30, 2.90, 2.50]      & [2.00, 1.88, 1.80]      \\
S1-A-High  &                     & A               & High                 &                                     & [4.71, 4.14, 3.57]      & [4.00, 3.75, 3.60]      \\
S1-B-Med   &                     & B               & Medium               &                                     & [3.30, 2.90, 2.50]      & [2.00, 1.88, 1.80]      \\
S1-C-Med   &                     & C               & Medium               &                                     & [3.30, 2.90, 2.50]      & [2.00, 1.88, 1.80]      \\
\midrule
S2-A-Low   & \multirow{5}{*}{S2} & A               & Low                  & \multirow{5}{*}{[0.23, 0.38, 0.49]} & [1.03, 0.91, 0.78]      & [3.20, 3.00, 2.88]      \\
S2-A-Med   &                     & A               & Medium               &                                     & [2.75, 2.42, 2.08]      & [5.00, 4.69, 4.50]      \\
S2-A-High  &                     & A               & High                 &                                     & [8.25, 7.25, 6.25]      & [8.89, 8.33, 8.00]      \\
S2-B-Med   &                     & B               & Medium               &                                     & [2.75, 2.42, 2.08]      & [5.00, 4.69, 4.50]      \\
S2-C-Med   &                     & C               & Medium               &                                     & [2.75, 2.42, 2.08]      & [5.00, 4.69, 4.50]      \\
\midrule
S3-A-Low   & \multirow{5}{*}{S3} & A               & Low                  & \multirow{5}{*}{[0.10, 0.65, 0.41]} & [2.12, 1.16, 1.92]      & [2.00, 1.88, 1.80]      \\
S3-A-Med   &                     & A               & Medium               &                                     & [3.79, 2.07, 3.43]      & [3.20, 3.00, 2.88]      \\
S3-A-High  &                     & A               & High                 &                                     & [4.44, 3.22, 4.44]      & [4.00, 3.75, 3.60]      \\
S3-B-Med   &                     & B               & Medium               &                                     & [3.79, 2.07, 3.43]      & [3.20, 3.00, 2.88]      \\
S3-C-Med   &                     & C               & Medium               &                                     & [3.79, 2.07, 3.43]      & [3.20, 3.00, 2.88]      \\
\bottomrule
\end{tabular}
\end{table*}

\subsection{Real-world Dataset}
\label{app:dataset_real}

\begin{figure}[H]
  \centering
  \includegraphics[width=\linewidth]{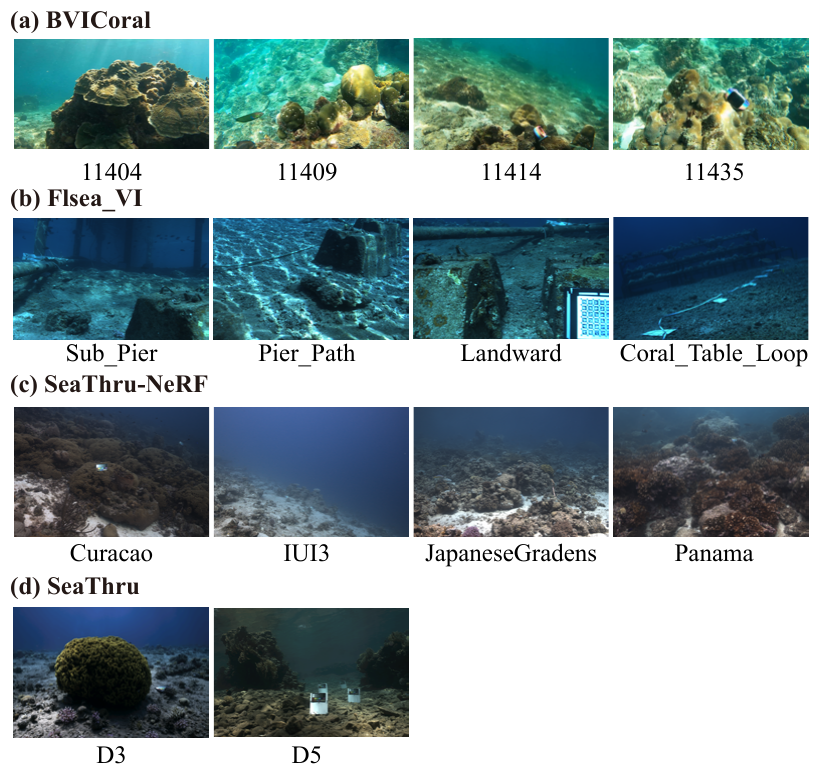}
  \caption{The real-world datasets used in our experiments.}
  \Description{Representative images from the real-world underwater datasets.}
  \label{fig:app_dataset_all}
\end{figure}

To evaluate generalization in real underwater environments, we use four public datasets covering diverse water types, lighting conditions, visibility levels, camera motions, and scene structures,  as illustrated in Fig.~\ref{fig:app_dataset_all}.

\textbf{BVICoral Dataset.}
The BVICoral dataset~\cite{anantrasirichai_2024_11093417} targets underwater 3D reconstruction with videos captured around coral reef structures. We select four sequences, 11404, 11409, 11414, and 11435, which contain strong attenuation, backscattering, and local flickering.

\textbf{Flsea\_VI Dataset.}
The Flsea\_VI dataset~\cite{randall2023flsea} provides underwater visual-inertial sequences collected in shallow waters of the Mediterranean and Red Sea. We use Sub\_Pier, Pier\_Path, Landward, and Coral\_Table\_Loop, which provide forward-looking motion, diverse objects, and complex lighting with strong caustics and flickering.

\textbf{SeaThru-NeRF Dataset.}
The SeaThru-NeRF dataset~\cite{levy2023seathru} contains four scenes, Curacao, IUI3, Japanese Gardens, and Panama, captured under varying water conditions. Each scene includes 18--29 undistorted images with known camera intrinsics and COLMAP poses, covering a wide range of visibility, depth, and scattering conditions.

\textbf{SeaThru Dataset.}
The SeaThru dataset~\cite{akkaynak2019sea} includes horizontally captured underwater scenes. We use D3 and D5, with 68 and 43 images, respectively. Both include calibrated color charts, enabling quantitative evaluation of color restoration.

\section{Additional Implementation Details}
\label{app:implementation}

\textbf{Water Parameters Extractor (WPE).}
The WPE takes as input a degraded image along with its depth-enhanced version, obtained via element-wise multiplication with the pseudo-depth map. These two inputs are concatenated along the channel dimension and passed through a sequence of three depthwise separable convolution (DW-Conv) blocks, each followed by ReLU activations and average pooling layers. The resulting feature maps are flattened and projected through a linear layer to regress the global water parameters: ambient light $\bm{A}_t$, attenuation coefficient $\bm{\beta}_t$, and backscatter coefficient $\bm{\gamma}_t$.

\textbf{Instantaneous Brightness Feature Encoder (IBF).}
The IBF encoder processes the degraded image using a series of convolutional blocks interleaved with ReLU activations and average pooling. Shortcut connections are employed to preserve critical features across stages. The module outputs two components: (1) a low-resolution spatial feature map $\bm{F}_l$ that captures localized lighting variations, and (2) a global feature vector $\bm{f}_g$ that encodes the overall scene brightness.

\textbf{Position Encoder $\phi(\cdot)$.}
The Position Encoder $\phi(\cdot)$ is implemented through a learnable hash-based encoder~\cite{muller2022instant}. Specifically, the 3D position $\bm{\mu}_i$ of each Gaussian is first normalized to $[0, 1]$ and then passed through a multi-level HashEncoding module with 16 levels, resolutions ranging from $16$ to $8192$, and a hash table size of $2^{21}$. Each level outputs 2 features, yielding a compact yet expressive position descriptor.

\textbf{Color Encoder $\omega(\cdot)$.}
The Color Encoder $\omega(\cdot)$ is a lightweight two-layer MLP with ReLU activations and a hidden width of 16. The output is a 32-dimensional color embedding, providing a learned feature representation for color-based modulation.

\textbf{Illumination Perturbation Decoder.}
The Illumination Perturbation Decoder is implemented as a multi-layer perceptron (MLP) consisting of ${L}=2$ hidden layers with a hidden width of 64 and ReLU activations. Its input is formed by concatenating four types of features: (1) a local brightness feature vector $\bm{f}_l \in \mathbb{R}^{16}$, (2) a global scene feature vector $\bm{f}_g \in \mathbb{R}^{16}$, (3) the position embedding $\phi(\bm{\mu}_i) \in \mathbb{R}^{32}$ generated by the hash-based encoder, and (4) the intrinsic color embedding $\omega(\bm{c}_i) \in \mathbb{R}^{32}$ obtained from a two-layer MLP. The concatenated input vector therefore lies in $\mathbb{R}^{96}$. This 96-dimensional vector is fed into the Illumination Perturbation Decoder, which outputs a 3-dimensional feature as the transient color offset $\epsilon_{it}$.

\textbf{Optimization Details.}
Our method builds upon the gsplat framework~\cite{ye2025gsplat}, with customized optimization schedules for the SDM module. We use the Adam optimizer with $\epsilon=10^{-15}$ across all components. The learning-rate schedules are defined over the 30{,}000-step training process. The Illumination Perturbation Decoder, Water Parameters Extractor, and intrinsic color encoder $\omega(\cdot)$ use an exponentially decaying learning rate from $1\times10^{-3}$ to $1.5\times10^{-4}$. The IBF encoder uses a slightly higher initial learning rate of $2\times10^{-3}$, decayed to $2\times10^{-4}$ with a 1024-step warm-up. The position encoder $\phi(\cdot)$ adopts a more aggressive schedule, with its learning rate decaying from $1\times10^{-2}$ to $1\times10^{-4}$.

\end{document}